\documentclass{article}
\usepackage{spconf,amsmath,graphicx,hyperref}
\usepackage{enumitem}
\usepackage{amssymb}
\usepackage{xcolor}
\usepackage{multirow}
\usepackage{booktabs}
\usepackage{xspace}
\usepackage{mathalpha}
\definecolor{l_color}{RGB}{0,0,0}
\definecolor{f_color}{RGB}{0,0,0}
\definecolor{baselinecolor}{gray}{.9}
\definecolor{demphcolor}{RGB}{144,144,144}
\newcommand{\demph}[1]{\textcolor{demphcolor}{#1}}
\usepackage[table]{xcolor}
\title{Towards Robust Visual Continual Learning with \\
Multi-Prototype Supervision}
\name{Xiwei Liu$^1$, Yulong Li$^1$, Yichen Li$^{1,2}$, Xinlin Zhuang$^{1,3}$, Haolin Yang$^1$, Huifa Li$^1$, Imran Razzak$^{1,\dagger}$\thanks{$^{\dagger}$ Corresponding author.}}
\address{$^{1}$Mohamed bin Zayed University of Artificial Intelligence\\
$^{2}$Huazhong University of Science and Technology, $^{3}$East China Normal University}
\begin{document}
%\ninept
%

\maketitle
%
% \normalsize
\begin{abstract}
Language-guided supervision, which utilizes a frozen semantic target from a Pretrained Language Model (PLM), has emerged as a promising paradigm for visual Continual Learning (CL). However, relying on a single target introduces two critical limitations: 1) semantic ambiguity, where a polysemous category name results in conflicting visual representations, and 2) intra-class visual diversity, where a single prototype fails to capture the rich variety of visual appearances within a class. To this end, we propose MuproCL, a novel framework that replaces the single target with multiple, context-aware prototypes. Specifically, we employ a lightweight LLM agent to perform category disambiguation and visual-modal expansion to generate a robust set of semantic prototypes. A LogSumExp aggregation mechanism allows the vision model to adaptively align with the most relevant prototype for a given image.
Extensive experiments across various CL baselines demonstrate that MuproCL consistently enhances performance and robustness, establishing a more effective path for language-guided continual learning.

\end{abstract}
\vspace{-0.2em}
\section{Introduction}
\label{sec:intro}
Continual learning (CL) equips machine learning systems with the crucial ability to acquire new knowledge sequentially without overwriting previous learning, a process known as mitigating \textit{catastrophic forgetting}~\cite{nguyen2019toward}. This capability is essential for dynamic, real-world applications such as robotics, healthcare, and autonomous driving~\cite{lee2020clinical, verwimp2023continual}. 
% To alleviate CF, researchers have explored various routes. 
Key approaches to alleviate catastrophic forgetting include regularization-based methods that constrain significant parameter changes~\cite{aljundi2018memory, chaudhry2018riemannian, zenke2017continual}; replay-based techniques that revisit past data, either stored or synthetically generated~\cite{aljundi2019gradient, ostapenko2019learning, shi2022mimicking}; distillation-based approaches that transfer knowledge from a previous model state~\cite{wu2019large, douillard2020podnet, tao2020few}; and methods that dynamically adapt the network architecture for new tasks~\cite{douillard2022dytox, li2019learn, yan2021dynamically}.

However, most existing methods rely on randomly initialized classifiers with one-hot supervision, thereby overlooking the rich semantic information embedded in category names. Recently, a promising paradigm has emerged to tackle this issue by leveraging knowledge from pretrained language models (PLMs)~\cite{dosovitskiy2020image}. LingoCL~\cite{ni2024enhancing} first novely introduces the use of a PLM to generate a single semantic vector for each class, which then serves as a frozen, static classifier. This language-guided supervision provides a stable and knowledge-rich learning target, proving highly effective in mitigating representation drifting and enhancing knowledge transfer.

While language-guided supervision is a promising solution for CL, the reliance on a single, context-free semantic target, as practiced by LingoCL, reveals two notable challenges. First, it struggles with semantic ambiguity. A single category name can correspond to multiple visually distinct concepts (e.g., "crane" as a bird versus construction equipment). A single prototype for such a polysemous word inevitably create a conflicting learning objective that forces the model to map visually disparate images to the same point in the feature space. This can distort the representations and impede learning. Second, this approach lacks robustness to intra-class visual diversity. Even for monosemous words, visual manifestations can be highly multi-modal (e.g., "apple" as a fruit versus a company logo; "peeled apple" versus "bunch of apples"). A single semantic target fails to cover this rich diversity, potentially leading to biases and suboptimal performance on less common visual modes~\cite{ma2025does}. These issues highlight the fragility of static semantic supervision in open-world settings.
\begin{figure*}[t]
\setlength{\abovecaptionskip}{2pt}
    \centering
    \includegraphics[width=\textwidth]{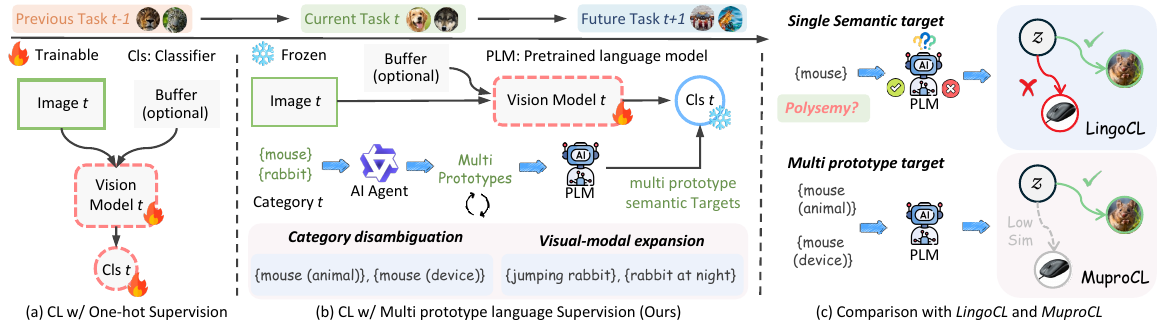}
    \caption{(a) Overview of the typical CL methods which are supervised only by one-hot labels, coupled with the randomly initialized classifier head, and the encoder and classifier head are jointly optimized. (b) Overview of our proposed MuproCL which is supervised by multi prototype semantic targets generated from the pretrained language model. (c) Motivational comparison in the feature space, where $z$ represents the output feature of an input image from vision encoder. Top: A single-target method like LingoCL risks forcing $z$  to align with an incorrect target due to semantic ambiguity (polysemy). Bottom: MuproCL provides multiple prototypes, enabling $z$  to adaptively align with the correct target via LogSumExp. Low Sim: Low similarity.} 
    \label{fig:overview}
\end{figure*}

In this paper, we propose MuproCL, a simple yet effective paradigm that addresses these issues. Instead of relying on a single ambiguous category name, MuproCL utilizes a lightweight AI agent to generate multiple, context-aware semantic prototypes for each class. Specifically, our approach performs (a) category disambiguation to create distinct prototypes for different meanings of polysemous words, and (b) visual-modal expansion to capture the varied appearances of monosemous words. These multi-prototype then form a robust, frozen classifier. By employing a LogSumExp~\cite{ramon2000multi} aggregation mechanism during training, the vision model can adaptively align image features with the most relevant semantic prototype. Extensive experiments demonstrate that MuproCL consistently enhances the performance and robustness of various CL baselines, paving the way for more effective applications of language-guided supervision.

\section{Method}
\label{sec:method}
\subsection{Revisiting Classifier in CL}
In typical CL scenarios, the vision encoder $g_V$ is sequentially trained over tasks. For $t$-th task, we denote the training dataset as $\mathcal{D}_t = \{(\boldsymbol{x}_t, y_t)\}$ that contains $C_t$ disjoint classes. A prevailing paradigm involves jointly training the vision encoder $g_V$ with a task-specific classifier $\mathbf{W}_t \in \mathbb{R}^{C_t\times d}$ for each new task. Conventionally, the classifier's weights, which serve as semantic targets, are initialized from a random initialized, e.g., $\mathbf{W}_t \sim \mathcal{N}(\mathbf{0}, \mathbf{I}_d)$. The model is optimized with the standard classification objective under stationary conditions:
\begin{equation}
    \textcolor{l_color}{g^*_V}, \textcolor{l_color}{\mathbf{W}_t^*} = \underset{\textcolor{l_color}{\Theta_V},\textcolor{l_color}{\mathbf{W}_t}}{\mathrm{arg min}}~
                {\mathbb{E}}_{{\boldsymbol{x}_t,y_t \sim \mathcal{D}_t}}\big[ \mathcal{L}(\mathrm{sim}(\textcolor{l_color}{\mathbf{W}_t}, \textcolor{l_color}{g_V}(\boldsymbol{x}_t)), y_t) \big], \label{eq:cl_paradigm}
\end{equation}
Here, the function $\mathrm{sim}(\mathbf{W}_t, g_V(\boldsymbol{x}_t))$ calculates the similarity between the image embedding $g_V(\boldsymbol{x})$ and every semantic target in \textcolor{black}{$\mathbf{W}_t$}, using the inner product. However, this approach's reliance on randomly initialized classifiers creates two fundamental bottlenecks in CL:\textit{ inefficient knowledge transfer} due to the absence of semantic priors, and \textit{representation drifting} from task-specific optimization conflicts~\cite{ni2024enhancing}.

To address these limitations, LingoCL~\cite{ni2024enhancing} recently pioneered replacing the trainable classifier $\mathbf{W}_t$ with a frozen, static one containing semantic knowledge from a PLM:
\begin{equation}
        \textcolor{l_color}{g^*_V} = \underset{\textcolor{l_color}{\Theta_V}}{\mathrm{arg min}}~
                    {\mathbb{E}}_{{\boldsymbol{x}_t,y_t \sim \mathcal{D}_t}}\big[ \mathcal{L}(\mathrm{sim}(\textcolor{f_color}{\tilde{\mathbf{W}}_t}, \textcolor{l_color}{g_V}(\boldsymbol{x}_t)), y_t) \big].
\end{equation}
While providing a knowledge-rich signal, this single-target paradigm introduces its own critical limitations. It struggles with semantic ambiguity, as a single category name with potential polysemy can correspond to visually disparate concepts, leading to conflicting learning objectives. 

\setlength{\tabcolsep}{1.0mm}{
\begin{table*}[t!]
  \centering
  % \label{tab:main}
  \resizebox{\textwidth}{!}{
  \begin{tabular}
  {p{2.5cm}p{2.5cm}p{1.5cm}p{1.5cm}p{1.3cm}p{0.cm}p{1.5cm}p{1.5cm}p{1.3cm}p{0.cm}p{1.5cm}p{1.5cm}p{1.3cm}}
  \toprule
   \multirow{3}{*}{Method} & \multirow{3}{*}{Backbone} & \multicolumn{3}{c}{\emph{$B$=10, $C$=10}} && \multicolumn{3}{c}{\emph{$B$=5, $C$=5}} && \multicolumn{3}{c}{\emph{$B$=2, $C$=2}} \\
  \cmidrule{3-5} \cmidrule{7-9} \cmidrule{11-13} 
   & & Avg ($\uparrow$) & Last ($\uparrow$) & $\mathcal{F}$ ($\downarrow$) && Avg ($\uparrow$) & Last ($\uparrow$) & $\mathcal{F}$ ($\downarrow$) &&  Avg ($\uparrow$) & Last ($\uparrow$) & $\mathcal{F}$ ($\downarrow$) \\
    \midrule
    Oracle  & ResNet-18 & 77.6 & 77.6 & - && 77.6 & 77.6 & - && 77.6 & 77.6 & -  \\
    \quad w/ LingoCL & & 78.0 & 78.0 & - && 78.0 & 78.0 & - && 78.0 & 78.0 & - \\
    \rowcolor[gray]{0.9}\quad w/ MuproCL & & 78.3 & 78.3 & - && 78.3 & 78.3 & - && 78.3 & 78.3 & - \\
    \midrule
    \multicolumn{5}{l}{\textit{Architecture-based methods}} \\
    AANet~\cite{liu2021adaptive} & ResNet-18  & 64.6\scriptsize{$\pm$0.2} & 49.1\scriptsize{$\pm$0.2} & - && 62.5\scriptsize{$\pm$0.3} & 42.5\scriptsize{$\pm$0.3} & - && 57.7\scriptsize{$\pm$0.5} & 37.6\scriptsize{$\pm$0.5} & - \\

    \quad w/ LingoCL && 65.4\scriptsize{$\pm$0.1}  & 50.4\scriptsize{$\pm$0.1}  & - && 63.2\scriptsize{$\pm$0.2}  & 44.5\scriptsize{$\pm$0.4}  & - && 58.7\scriptsize{$\pm$0.4}  & 38.6\scriptsize{$\pm$0.5}  & -  \\
    \rowcolor[gray]{0.9}\quad w/ MuproCL && \textbf{65.9}\scriptsize{$\pm$0.1}  & \textbf{51.1}\scriptsize{$\pm$0.2}  & -  && \textbf{63.9}\scriptsize{$\pm$0.2}  & \textbf{45.4}\scriptsize{$\pm$0.4}  & -  && \textbf{60.0}\scriptsize{$\pm$0.1}  & \textbf{39.4}\scriptsize{$\pm$0.3}  & - \\

    DyTox~\cite{douillard2022dytox} & ConViT & 69.5\scriptsize{$\pm$0.0} & 52.8\scriptsize{$\pm$0.2} & 33.0\scriptsize{$\pm$0.0} && 67.4\scriptsize{$\pm$0.1} & 48.1\scriptsize{$\pm$0.3} & 37.8\scriptsize{$\pm$0.0} && 64.5\scriptsize{$\pm$0.2} & 44.8\scriptsize{$\pm$0.3} & 41.3\scriptsize{$\pm$0.1} \\
    \quad w/ LingoCL && 71.9\scriptsize{$\pm$0.0}  & 58.9\scriptsize{$\pm$0.1}  & 24.9\scriptsize{$\pm$0.0}  && 70.0\scriptsize{$\pm$0.1}  & 52.3\scriptsize{$\pm$0.3}  & 30.5\scriptsize{$\pm$0.1}  && 65.9\scriptsize{$\pm$0.1}  & 46.3\scriptsize{$\pm$0.3}  & 36.1\scriptsize{$\pm$0.2} \\
    \rowcolor[gray]{0.9}\quad w/ MuproCL && \textbf{72.7}\scriptsize{$\pm$0.1}  & \textbf{61.3}\scriptsize{$\pm$0.2}  & \textbf{22.1}\scriptsize{$\pm$0.1}  && \textbf{71.5}\scriptsize{$\pm$0.1}  & \textbf{53.1}\scriptsize{$\pm$0.2}  & \textbf{28.1}\scriptsize{$\pm$0.1}  && \textbf{67.2}\scriptsize{$\pm$0.1}  & \textbf{47.7}\scriptsize{$\pm$0.2}  & \textbf{34.3}\scriptsize{$\pm$0.2} \\
    
    \midrule
    \multicolumn{5}{l}{\textit{Distillation-based methods}} \\
    LUCIR~\cite{hou2019learning} & ResNet-18  & 60.2\scriptsize{$\pm$0.4} & 46.5\scriptsize{$\pm$0.7} & 37.3\scriptsize{$\pm$0.5} && 54.8\scriptsize{$\pm$0.7} & 41.7\scriptsize{$\pm$1.0} & 42.0\scriptsize{$\pm$0.5} && 45.6\scriptsize{$\pm$1.1} & 36.2\scriptsize{$\pm$1.6} & 44.5\scriptsize{$\pm$0.9} \\
    \quad  w/ LingoCL && 61.9\scriptsize{$\pm$0.3}  & 47.5\scriptsize{$\pm$0.6}  & 36.5\scriptsize{$\pm$0.1} && 56.3\scriptsize{$\pm$0.5}  & 44.3\scriptsize{$\pm$0.4}  & \textbf{39.8}\scriptsize{$\pm$0.8}  && 46.8\scriptsize{$\pm$1.2}  & 37.0\scriptsize{$\pm$0.6}  & 42.3\scriptsize{$\pm$1.0}  \\
    \rowcolor[gray]{0.9}\quad w/ MuproCL && \textbf{62.6}\scriptsize{$\pm0.5$}  & \textbf{48.3}\scriptsize{$\pm0.6$}  & \textbf{35.6}\scriptsize{$\pm0.3$}  && \textbf{56.8}\scriptsize{$\pm0.4$}  & \textbf{45.0}\scriptsize{$\pm0.8$}  & \textbf{38.6}\scriptsize{$\pm0.9$}  && \textbf{47.2}\scriptsize{$\pm1.1$}  & \textbf{38.3}\scriptsize{$\pm1.3$}  & \textbf{41.1}\scriptsize{$\pm0.9$} \\

    BiC~\cite{wu2019large} & ResNet-18 & 57.8\scriptsize{$\pm0.9$} & 41.2\scriptsize{$\pm1.0$} & 26.7\scriptsize{$\pm1.0$} && 50.1\scriptsize{$\pm0.6$} & 34.7\scriptsize{$\pm0.1$} & 28.7\scriptsize{$\pm0.7$} && 38.1\scriptsize{$\pm1.0$} & 23.6\scriptsize{$\pm0.4$} & 38.6\scriptsize{$\pm0.9$} \\ 
    \quad w/ LingoCL && 60.1\scriptsize{$\pm0.5$}  & 43.4\scriptsize{$\pm1.0$}  & 20.5\scriptsize{$\pm1.0$}  && 51.6\scriptsize{$\pm0.6$}  & 36.6\scriptsize{$\pm1.3$}  & 19.6\scriptsize{$\pm0.6$}  && 44.9\scriptsize{$\pm1.0$}  & 31.1\scriptsize{$\pm0.8$}  & 29.7\scriptsize{$\pm0.6$} \\
    \rowcolor[gray]{0.9}\quad w/ MuproCL && \textbf{61.2}\scriptsize{$\pm0.6$}  & \textbf{44.1}\scriptsize{$\pm0.8$}  & \textbf{20.2}\scriptsize{$\pm0.5$}  && \textbf{51.9}\scriptsize{$\pm0.5$}  & \textbf{37.1}\scriptsize{$\pm0.6$}  & \textbf{19.5}\scriptsize{$\pm0.7$}  && \textbf{46.8}\scriptsize{$\pm0.9$}  & \textbf{33.4}\scriptsize{$\pm0.7$}  & \textbf{28.5}\scriptsize{$\pm0.6$} \\

    \midrule
    \multicolumn{5}{l}{\textit{Rectification-based methods}} \\
    
    CwD~\cite{shi2022mimicking} & ResNet-18 & 60.0\scriptsize{$\pm0.5$} & 46.7\scriptsize{$\pm0.5$} & 36.9\scriptsize{$\pm0.9$} && 54.4\scriptsize{$\pm0.6$} & 42.2\scriptsize{$\pm0.5$} & 41.5\scriptsize{$\pm0.4$} && 40.2\scriptsize{$\pm1.2$} & 34.0\scriptsize{$\pm1.2$} & 44.6\scriptsize{$\pm0.2$} \\
    \quad w/ LingoCL && 60.6\scriptsize{$\pm0.7$}  & 47.6\scriptsize{$\pm1.1$}  & 35.3\scriptsize{$\pm1.0$}  && 55.7\scriptsize{$\pm0.9$}  & 44.3\scriptsize{$\pm0.5$} & 39.2\scriptsize{$\pm0.7$}  && 46.0\scriptsize{$\pm0.7$}  & 38.4\scriptsize{$\pm0.8$}  & 36.9\scriptsize{$\pm0.8$}  \\
    \rowcolor[gray]{0.9}\quad w/ MuproCL && \textbf{60.9}\scriptsize{$\pm0.4$}  & \textbf{48.0}\scriptsize{$\pm0.8$}  & \textbf{34.7}\scriptsize{$\pm1.0$}  && \textbf{55.8}\scriptsize{$\pm0.3$}  & \textbf{44.7}\scriptsize{$\pm0.5$}  & \textbf{38.5}\scriptsize{$\pm0.7$}  && \textbf{48.2}\scriptsize{$\pm0.9$}  & \textbf{40.6}\scriptsize{$\pm0.8$}  & \textbf{36.1}\scriptsize{$\pm0.7$} \\
    
    IL2M~\cite{belouadah2019il2m} & ResNet-18  & 57.8\scriptsize{$\pm0.3$} & 44.3\scriptsize{$\pm0.8$} & 41.0\scriptsize{$\pm0.3$} && 52.6\scriptsize{$\pm0.8$} & 40.5\scriptsize{$\pm0.4$} & 45.3\scriptsize{$\pm1.0$} && 44.0\scriptsize{$\pm1.1$} & 34.2\scriptsize{$\pm0.8$} & 48.5\scriptsize{$\pm0.5$} \\
    \quad w/ LingoCL && 62.1\scriptsize{$\pm0.0$}  & 48.1\scriptsize{$\pm1.0$}  & 37.8\scriptsize{$\pm0.3$}  && 56.6\scriptsize{$\pm0.4$}  & 44.0\scriptsize{$\pm1.0$}  & 43.0\scriptsize{$\pm1.2$}  && 48.0\scriptsize{$\pm0.7$}  & 39.6\scriptsize{$\pm0.0$}  & 42.8\scriptsize{$\pm0.2$}  \\
    \rowcolor[gray]{0.9}\quad w/ MuproCL && \textbf{63.4}\scriptsize{$\pm0.1$}  & \textbf{49.7}\scriptsize{$\pm0.9$}  & \textbf{36.3}\scriptsize{$\pm0.3$}  && \textbf{58.5}\scriptsize{$\pm0.5$}  & \textbf{46.9}\scriptsize{$\pm1.0$}  & \textbf{41.7}\scriptsize{$\pm1.1$}  && \textbf{49.5}\scriptsize{$\pm0.9$}  & \textbf{40.8}\scriptsize{$\pm0.5$}  & \textbf{40.9}\scriptsize{$\pm0.3$} \\
    
  \bottomrule
\end{tabular}
}
\caption{Results on class incremental experiments on CIFAR100 of Average accuracy (\%), last phase accuracy (\%) and forgetting rate $\mathcal{F}$ (\%) with and without text-supervised classifier at various CL settings. $B$ denotes the number of classes at the initial task, and $C$ denotes the number of classes in each task after the initial one. Notably, for each metric, $\uparrow$ ($\downarrow$) indicates that the larger (the smaller) values, the better results are. 'Oracle' represents training the model on data from all tasks at once.}
\label{tab:main}
 \vspace{-0.2cm}
\end{table*}
}
\subsection{Multi-Prototype Supervision}
To better address category names polysemy and expand visual-mode diversity while preserving the benefits of a frozen classifier in CL, we utilize a light-weight multi-prototype generation and aggregation, as illustrated in Figure~\ref{fig:overview}. Given an incoming task $t$, the procedure is as follows: 

(I) Gathering the category names $[\boldsymbol{l}_1,\cdots,\boldsymbol{l}_{C_t}]$ of task $t$.

(II) For each category $\boldsymbol{l}_c$, an LLM-based agent $\mathcal{A}$ produces a small set of complementary textual prompts $\mathcal{S}_c$ by (a) \textbf{polysemy detection and disambiguation} if $\boldsymbol{l}_c$ is polysemous (e.g., \texttt{crane(bird)}, \texttt{crane(construction equipment)}), or (b) \textbf{visual-mode expansion} if $\boldsymbol{l}_c$ is monosemous (e.g., view/style/context such as ``$l_c$ at night'', ``a logo of $l_c$'', ``a sketch of $l_c$''). In practice, we instantiate $\mathcal{A}$ as a filter--select pipeline: an LLM first proposes a pool of candidate prompts; we embed all candidates with the same PLM $g_T$ used for the classifier and then (i) discard non-visual or non-denotational senses, (ii) merge or remove near-duplicates whose pairwise cosine exceeds $0.95$, and (iii) select a diversity-preserving subset by farthest-point sampling until the marginal coverage gain falls below a threshold 0.2, thereby adapting $K_c$ per class capped at $K_{\max}=4$.
\begin{equation}
    \mathcal{S}_c \;=\; \{\, \boldsymbol{l}_c^{(k)} \,\}_{k=1}^{K_c}, \qquad 1 \le K_c \le K_{\max},
\end{equation}

(III) Feeding the generated prompt subsets into the PLM to produce multi semantic targets for the classifier:
\begin{equation}
    \textcolor{f_color}{\tilde{\mathbf{W}}_t} = \textcolor{f_color}{g_T}([\mathcal{S}_1,\cdots,\mathcal{S}_{C_t}])\in\mathbb{R}^{\sum_{c=1}^{C_t}K_c\times d},
    \end{equation}
    
(IV) Optimizing the vision encoder \textcolor{l_color}{$g_V$} with LogSumExp aggregation, while keeping the \textcolor{f_color}{$\tilde{\mathbf{W}}_t$} \textit{frozen}. For convenience, we denote the $k$-th prototype of class $c$ by $\tilde w_c^{(k)}$. Given an input $x_t$ with feature $z=g_V(x_t)$, the score for class $c$ is:
\begin{equation}
\label{eq:smp-agg-vanilla}
s_c(x_t)\;=\;\log\!\sum_{k=1}^{K_c}\exp\!\Big(\mathrm{sim}\!\big(\tilde w_c^{(k)},\, z\big)\Big),
\end{equation}
Let $s(x_t)=[s_1(x_t),\ldots,s_{C_t}(x_t)]$. The encoder is optimized:
\begin{equation}
\label{eq:smp-train-vanilla}
g_V^{*}\;=\;\arg\min_{\Theta_V}\ \mathbb{E}_{(x_t,y_t)\sim D_t}\Big[\, \mathcal{L}\big(s(x_t),\, y_t\big)\, \Big].
\end{equation}

\section{Experiments}
\label{sec:Experiments}
\subsection{Experimental setup}
\textbf{Datasets.} We evaluate our method on CIFAR100~\cite{krizhevsky2009learning} under standard class-incremental learning (CIL) protocols. Let $B$ denote the number of classes in the initial task and $C$ the number of classes per subsequent task. We adopt $C=B$ to ensure each task contains an equal number of classes throughout the learning sequence. The memory buffer is fixed to 20 exemplars per class across all settings. All approaches are evaluated under the same class order for fair comparison.

\noindent\textbf{Metrics.} Following~\cite{ni2024enhancing}, MuproCL is extensively evaluated by three metrics: last-step accuracy (Last), average incremental accuracy (Avg), and forgetting rate ($\mathcal{F}$).

\noindent\textbf{Baselines.} We comprehensively evaluate MuproCL on six representative continual learning baselines, spanning various continual learning approaches. These include architecture-based methods like AANet~\cite{liu2021adaptive} and DyTox~\cite{douillard2022dytox}, distillation-based methods such as LUCIR~\cite{hou2019learning} and BiC~\cite{wu2019large}, and rectification-based methods like CwD~\cite{shi2022mimicking} and IL2M~\cite{belouadah2019il2m}.
\subsection{Implementation details}
To ensure the fair comparison, we adhered strictly to the officially released code and original hyperparameters for all baseline methods. 
All CNN-based models~\cite{liu2021adaptive, hou2019learning, wu2019large, shi2022mimicking, belouadah2019il2m} were trained for 160 epochs per task using an SGD~\cite{robbins1951stochastic} optimizer with a 0.1 initial learning rate, 0.9 momentum, and a batch size of 128. The learning rate was decreased by a factor of 10 at epochs 80 and 120 to ensure stable convergence. The ViT-based DyTox~\cite{douillard2022dytox} model was trained for 500 epochs per task using the Adam~\cite{kingma2014adam} optimizer with a 5e-4 learning rate and a 5-epoch warmup. After each task (except the first), DyTox was finetuned for 20 epochs on a balanced dataset with a 5e-5 learning rate.
% For CNN-based methods~\cite{liu2021adaptive, hou2019learning, wu2019large, shi2022mimicking, belouadah2019il2m}, we employ the SGD optimizer with an initial learning rate of 0.1, a momentum of 0.9, and a batch size of 128. Each model was trained for 160 epochs per task, with the learning rate decreased by a factor of 10 at the 80-th and 120-th epochs. For ViT-based methods DyTox~\cite{douillard2022dytox}, we train the model for 500 epochs per task using Adam~\cite{kingma2014adam} with a learning rate of 5e-4, including 5 epochs of warmup. At the end of each task (except the first), we finetune the model for 20 epochs with a learning rate of 5e-5 on a balanced dataset.
For the language supervision component, we utilized the Qwen2-7B-Instruct model~\cite{team2024qwen2} as the LLM agent. As for the PLM, we utilize the text transformer from CLIP-B/32~\cite{radford2021learning} pretrained on WIT-400M~\cite{radford2021learning}.

\subsection{Main Result}
Table~\ref{tab:main} presents the quantitative results on class incremental experiments on CIFAR100 by incorporating MuproCL into six distinct baselines. Across all evaluated continual learning settings, MuproCL consistently outperforms both the original baselines and the single semantic target LingoCL approach. For example, when applied to IL2M under the $B=10, C=10$ setting, our method improves the average accuracy by 5.6\% over the vanilla baseline and 1.3\% over LingoCL. Like LingoCL, MuproCL has only negligible impact on the performance of the Oracle model, implying that the observed gains stem primarily from mitigating forgetting rather than boosting single-task learning. This is further evidenced by the reduction in the forgetting rate. For instance, MuproCL reduces DyTox's forgetting rate by a remarkable 10.9\%.

\begin{figure} 
\centering
    \includegraphics[width=0.48\textwidth]{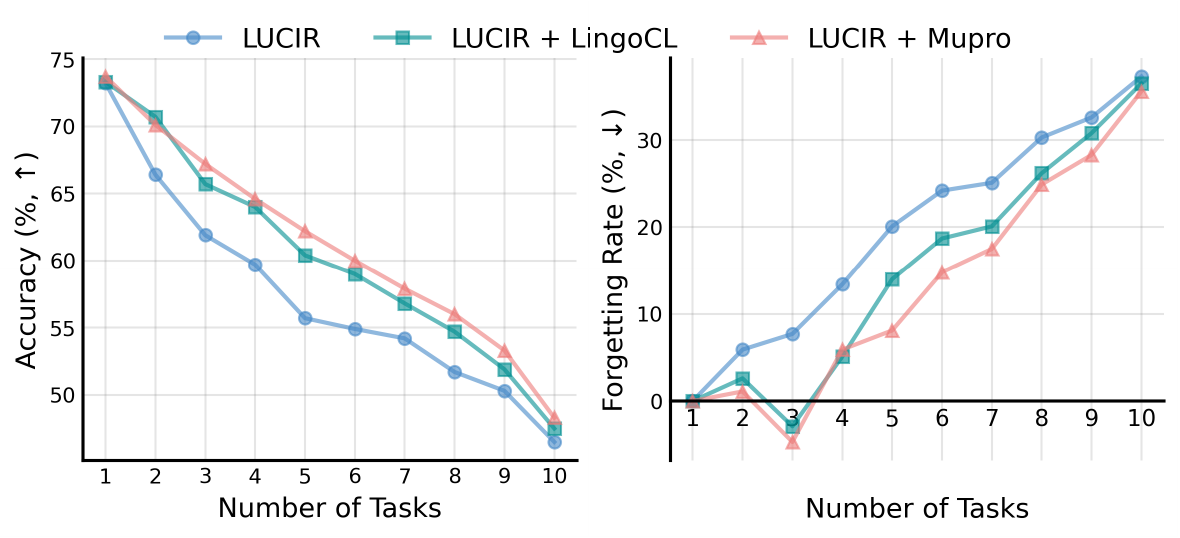}
    \caption{The evolution curve of accuracy and forgetting rate for each task under class-incremental setting $B=10, C=10$.} 
    \vspace{-0.3cm}
    \label{fig:curve}
\end{figure}

A key observation is that the benefits of MuproCL are amplified in more challenging and longer-sequence CL scenarios. Taking AANet as an example, the accuracy improvement grows from +1.3\% in the 10-task ($B=10, C=10$) setting to +2.3\% in the 50-task ($B=2, C=2$) setting. Such a result highlights the value of our multi-prototype supervision in mitigating the cumulative error and representational drift that accrue over extended learning horizons. As illustrated in Figure~\ref{fig:curve}, MuproCL maintains a consistent performance advantage over LingoCL, while also achieving a further reduction in forgetting. These results demonstrate that MuproCL provides both stronger accuracy and greater stability in CL.
% our method maintains a consistent, albeit slight, accuracy advantage over LingoCL across the incremental tasks while also further mitigating the overall forgetting rate.

\subsection{Ablation study}
\textbf{Prompt Template.} The formulation of the prompt is a critical factor of transferring knowledge from pretrained language models~\cite{radford2021learning}.
% Prompting~\cite{radford2021learning} is a widely used technique to transfer knowledge from pretrained language models. 
In Table~\ref{tab:abl 1}, we dissects the contributions of various prompting strategies. In setting \#2, we follow LingoCL by directly using the category name \texttt{\{object\}} as input without any additional templates. In setting \#3, we adopt the template \texttt{a photo of a \{object\}}. In setting \#4~\cite{radford2021learning}, we use an ensemble of 80 templates and average the resulting features. Our results show that the use of templates can slightly ease the forgetting, which we attribute to the improved stability of features produced by template ensembles. To validate our design, settings \#5 and \#6 ablate our model by disabling category disambiguation and visual-modal expansion, respectively. Both modifications lead to performance drops relative to the full model (\#7), confirming the necessity of each component. Nevertheless, even these reduced variants still outperform all single-prototype baselines, demonstrating the fundamental advantage of multi-prototype supervision.
\begin{table}[h]
  \centering
  \resizebox{0.5\textwidth}{!}{
  {
    \scriptsize
  \begin{tabular}{p{0.3cm}|lccc}
  \toprule
   \# & Template & Avg ($\uparrow$) & Last ($\uparrow$) & $\mathcal{F}$ ($\downarrow$) \\
   \midrule
   \demph{1} & \demph{Baseline-LUCIR}~\cite{hou2019learning} & \demph{60.2} & \demph{46.5} & \demph{37.3} \\
    \midrule
    2 & \texttt{\{object\}} & 61.9 & 47.5 & 36.5 \\
    3 & \texttt{a photo of a \{object\}} & 61.9 & 47.7 & 36.5 \\
    4 & Templates ensemble~\cite{radford2021learning} & 61.6 & 47.8 & \textbf{35.1} \\
    5 & MuproCL (w/o Disambiguation) & 62.0 & 47.7 & 36.3 \\
    6 & MuproCL (w/o Expansion) & 62.2 & 47.9 & 36.3 \\
    7 & MuproCL & \textbf{62.6} & \textbf{48.3} & 35.6 \\
  \bottomrule
\end{tabular}
  }
}
\caption{Comparison with different prompting techniques on CIFAR-100 under class-incremental setting $B=10, C=10$.}
\label{tab:abl 1}
\end{table}

\noindent\textbf{Effect of Prototypes Number.} We further investigate the impact of the maximum number of prototypes per class $K_{\max}$ on model performance based on LUCIR baseline.  As shown in Table~\ref{tab:abl 2}, the results reveal a clear trend. Increasing $K_{\max}$ from 1 to 4 leads to a steady improvement across all metrics. This confirms the fundamental benefit of our multi-prototype design. $K_{\max}$=8 achieve suboptimal performance. However, when further increasing the number to $K_{\max}$=16, we observe a slight performance degradation. This suggests that while multiple prototypes are crucial for covering semantic and visual diversity, an excessive number may introduce redundant or noisy signals that can hinder the learning process.
\begin{table}[h]
  \centering
  % \resizebox{0.5\textwidth}{!}{
  {
    % \scriptsize
  \begin{tabular}{lccc}
  \toprule
    $K_{\max}$ & Avg ($\uparrow$) & Last ($\uparrow$) & $\mathcal{F}$ ($\downarrow$) \\
    \midrule
     1 & 61.9\scriptsize{$\pm0.3$} & 47.5\scriptsize{$\pm0.6$}  & 36.5\scriptsize{$\pm0.1$}  \\
     2 & 62.4\scriptsize{$\pm0.4$}  & 48.1\scriptsize{$\pm0.4$}  & 36.0\scriptsize{$\pm0.3$}  \\
     4 & 62.6\scriptsize{$\pm0.5$}  & 48.3\scriptsize{$\pm0.6$}  & 35.6\scriptsize{$\pm0.3$}  \\
     8 & 62.5\scriptsize{$\pm0.2$} & 48.2\scriptsize{$\pm0.5$} & 35.6\scriptsize{$\pm0.4$} \\
     16 & 61.4\scriptsize{$\pm0.5$}  & 47.7\scriptsize{$\pm0.7$}  & 35.3\scriptsize{$\pm0.2$} \\
  \bottomrule
\end{tabular}
  % }
}
\caption{Effect of the maximum number of prototypes ($K_{\max}$) on CIFAR-100 under the $B=10, C=10$ setting.}
\label{tab:abl 2}
\end{table}
\vspace{-1em}
\section{CONCLUSION}
\label{sec:conclusion}
In this work, we address the critical limitations of single-target language supervision in CL. We introduce MuproCL, a novel paradigm that replaces the single, static target with multiple, context-aware semantic prototypes. By employing this robust, frozen multi-prototype classifier, MuproCL consistently enhances the performance of various CL baselines, establishing a more robust path for language-guided CL.

\vfill\pagebreak

\newpage
% \footnotesize
\ninept
\bibliographystyle{IEEEbib}
\bibliography{strings,refs}

\end{document}